\documentclass[journal]{IEEEtran}

\usepackage{graphicx}
\usepackage{multirow}
\usepackage{amsmath,amssymb,amsfonts}
\usepackage{amsthm}
\usepackage{mathrsfs}
\usepackage{xcolor}
\usepackage{textcomp}
\usepackage{manyfoot}
\usepackage{booktabs}
\usepackage{algorithm}
\usepackage{algpseudocode}
\usepackage{listings}
\usepackage{multirow}
\usepackage{tabularx}
\usepackage{cleveref}
\usepackage{comment}
\usepackage{stfloats}
\usepackage{url}
\usepackage{tikz}

\newcommand{\eg}{\textit{e}.\textit{g}.,\ }
\newcommand{\ie}{\textit{i}.\textit{e}.,\ }

\DeclareRobustCommand{\ss}[1]{\resizebox{!}{1.2ex}{#1}}

\begin{document}

\title{Lean Unet:\\ A Compact Model for Image Segmentation}

\author{\thanks{Funding provided partially by Swedish Research Council grant 2024-05664.}
\IEEEauthorblockN{Ture Hassler$^{1,2}$,
Ida Åkerholm$^{1}$,
Marcus Nordström$^{2}$,
Gabriele Balletti$^{2}$,
Orcun Goksel$^{1}$}\\[.5ex]
\IEEEauthorblockA{$^{1}$Department of Information Technology, Uppsala University, Uppsala, Sweden}\\
\IEEEauthorblockA{$^{2}$RaySearch Laboratories, Stockholm, Sweden}
}

\maketitle
\begin{abstract}
Unet and its variations have been standard in semantic image segmentation, especially for computer assisted radiology.
Current Unet architectures iteratively downsample spatial resolution while increasing channel dimensions to preserve information content.
Such a structure demands a large memory footprint, limiting training batch sizes and increasing inference latency.
Channel pruning compresses Unet architecture without accuracy loss, but requires lengthy optimization and may not generalize across tasks and datasets.
By investigating Unet pruning, we hypothesize that the final structure is the crucial factor, not the channel selection strategy of pruning.
Based on our observations, we propose a lean Unet architecture (LUnet) with a compact, flat hierarchy where channels are not doubled as resolution is halved.
We evaluate on a public MRI dataset allowing comparable reporting, as well as on two internal CT datasets. We show that a state-of-the-art pruning solution (STAMP) mainly prunes from the layers with the highest number of channels.
Comparatively, simply eliminating a random channel at the pruning-identified layer or at the largest layer achieves similar or better performance. Our proposed LUnet with fixed architectures and over 30 times fewer parameters achieves performance comparable to both conventional Unet counterparts and data-adaptively pruned networks.
The proposed lean Unet with constant channel count across layers requires far fewer parameters while achieving performance superior to standard Unet for the same total number of parameters.
Skip connections allow Unet bottleneck channels to be largely reduced, unlike standard encoder-decoder architectures requiring increased bottleneck channels for information propagation.
\end{abstract}

\section{Introduction}
Semantic image segmentation is a fundamental processing step in many medical applications including diagnosis, surgical planning, and intervention guidance -- all requiring automatic, accurate, and efficient methods.
Unet~\cite{unet_paper} is a neural network (NN) solution which has become the \mbox{de\,facto} standard for segmentation in most supervised learning settings.
It is also commonly used as a backbone in other NN settings such as for contrastive learning~\cite{Gomariz_joint_25}, detection/density estimation~\cite{Gomariz_probabilistic_22}, diffusion models~\cite{Webber_diffusion_24}, and generative adversarial networks~\cite{Zhang_learning_21} especially for processing medical images, where context is crucial.
In contrast to natural images, where observed objects are rarely attached to their background, imaging of anatomy is strictly context dependent, \ie the location and morphology of organs are relatively fixed.
Unet incorporates such contextual information with its multi-level-of-detail approach, by distilling contextual information at coarsening NN layers of its encoders, while employing such information for refining resolution in its decoder layers.
In contrast to conventional encoder-decoder architectures, skip connections in Unet facilitate the merging of contextual and resolution-specific information for precise pixel-resolved outputs.

Traditionally, deep learning has relied on the manual design of carefully constructed and empirically validated neural network architectures, often employing over-parameterized models to capture all potentially relevant information.
To avoid overfitting, one then requires large datasets as well as several optimization and regularization tricks such as Monte-Carlo dropouts at training time, optimized hyper-parameterizations, early-stopping, elastic weight regularization, among others.
From a function approximation point-of-view, by Occam's razor, and in practice, a network should be \emph{as complex as needed while as simple as possible} to provide an efficient, low-footprint representation that also prevents overfitting.
Neural Architecture Search (NAS)~\cite{White_neural_23} is an automated machine learning (AutoML) approach for finding optimal NN architectures by iterating over possible options using often black-box optimization approaches such as evolutionary search, reinforcement learning, and Bayesian function approximation.
Pruning is an alternative technique aiming at network compression by reducing NN model sizes via removing neurons or parameters that are selected based on various criteria~\cite{state_of_neural_network_pruning_blalock2020state}.
For Unet in particular, a recent method~\cite{stamp_paper} proposes iteratively pruning the neurons (\ie convolutions / channels) that produce the lowest activations during training.
This was shown to generate much smaller (sub)networks of initial Unet, while also reducing overfitting, which is a challenge especially for smaller training sets.

In this work, we closely examine gradual channel pruning in Unet, focusing particularly on the temporal evolution of channel selection during pruning.
We experimentally study the relative importance of the chosen specific channels and their locations in the architecture, with key comparisons to random and systematic (hand-crafted) pruning baselines, while interpreting these in the context of Lottery Ticket Hypothesis (LTH)~\cite{lottery_ticket_hyp_frankle2018lottery} and synaptic saliency~\cite{SynFlow}.
From our experiments, we conclude that such Unet pruning optimizes its architecture, rather than choosing channels with specific weights/activations -- an observation that corroborates some literature while contrasting with others.
Inspired by the asymptotic convergence of the NN architectures in our experiments, we propose a Lean Unet (LUnet) architecture with a fixed number of channels per block.
We show LUnet to perform comparably to or better than both pruned networks, which require complex, data-dependent processes, as well as corresponding Unet baselines, despite having over 30 times fewer parameters. 

\section{Methods}

\subsection{Neural Network Compression}
Network pruning can yield models capable of performing on par with, or superiorly to, their dense counterparts~\cite{lottery_ticket_hyp_frankle2018lottery, state_of_neural_network_pruning_blalock2020state} while using less than half of the original parameters, highlighting the substantial redundancy present in state-of-the-art deep neural networks. 

Some approaches aim for one-shot pruning prior to training, to remove structurally unimportant connections; for instance, SNIP~\cite{snip} proposed a saliency criterion based on gradient and weight magnitudes at the initial training step.

The Lottery Ticket Hypothesis (LTH)~\cite{lottery_ticket_hyp_frankle2018lottery} claims that certain subnetworks may perform particularly well due to their fortunate random initialization.
To utilize this, a NN is trained to convergence, the weights with the smallest magnitudes are removed, and the pruned subnetwork is retrained with its \emph{original} initial weights -- in a loop repeated iteratively until a desired sparsity is reached. 
LTH was later challenged by Liu \emph{et al.}~\cite{Rethinking_the_value_of_nn_pruning_Liu19}, who showed that an arbitrary initialization may generate similar results, indicating that the key factor is the architecture and not the initialized weights. 

Synaptic saliency is defined in~\cite{SynFlow} as a group of pruning criteria $\frac{\delta R}{\delta \Theta}*\Theta$ where $\Theta$ is the network parameters and $R$ is a function of the network output such as loss. 
Having demonstrated the conservation of cumulative synaptic saliency between the input and output of a layer, the authors argue that the individual neurons of a wider layer should have lower saliency on average.
Their method (SynFlow) hence is more likely to preferentially prune from larger layers.

\begin{figure}
\centering
\setlength{\tabcolsep}{1pt}
\renewcommand{\arraystretch}{0}
\begin{tabular}{@{}c@{}}
\includegraphics[height=.2\linewidth,trim=400 180 380 180,clip]{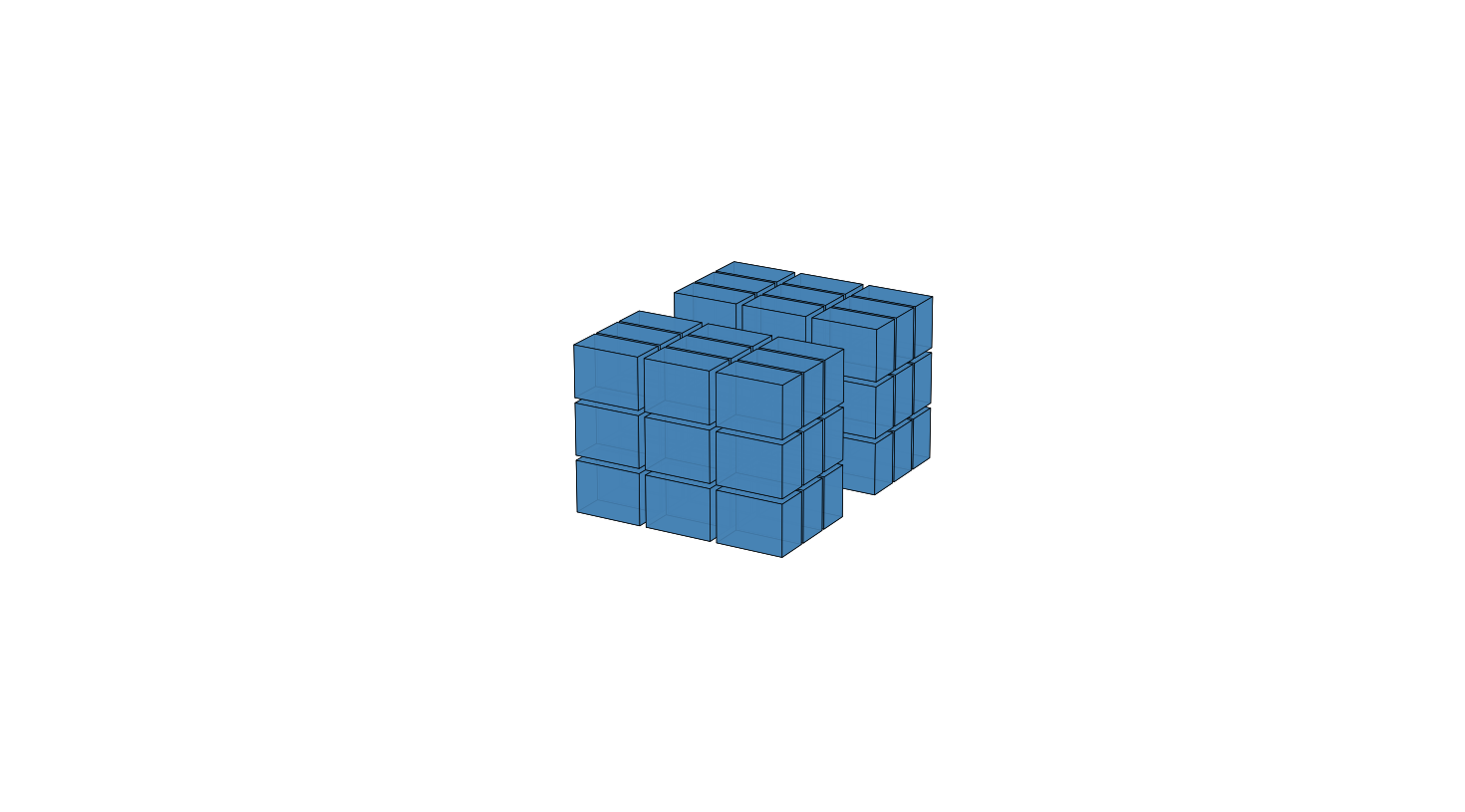}\\[-1mm]
\includegraphics[height=.2\linewidth,trim=400 180 380 180,clip]{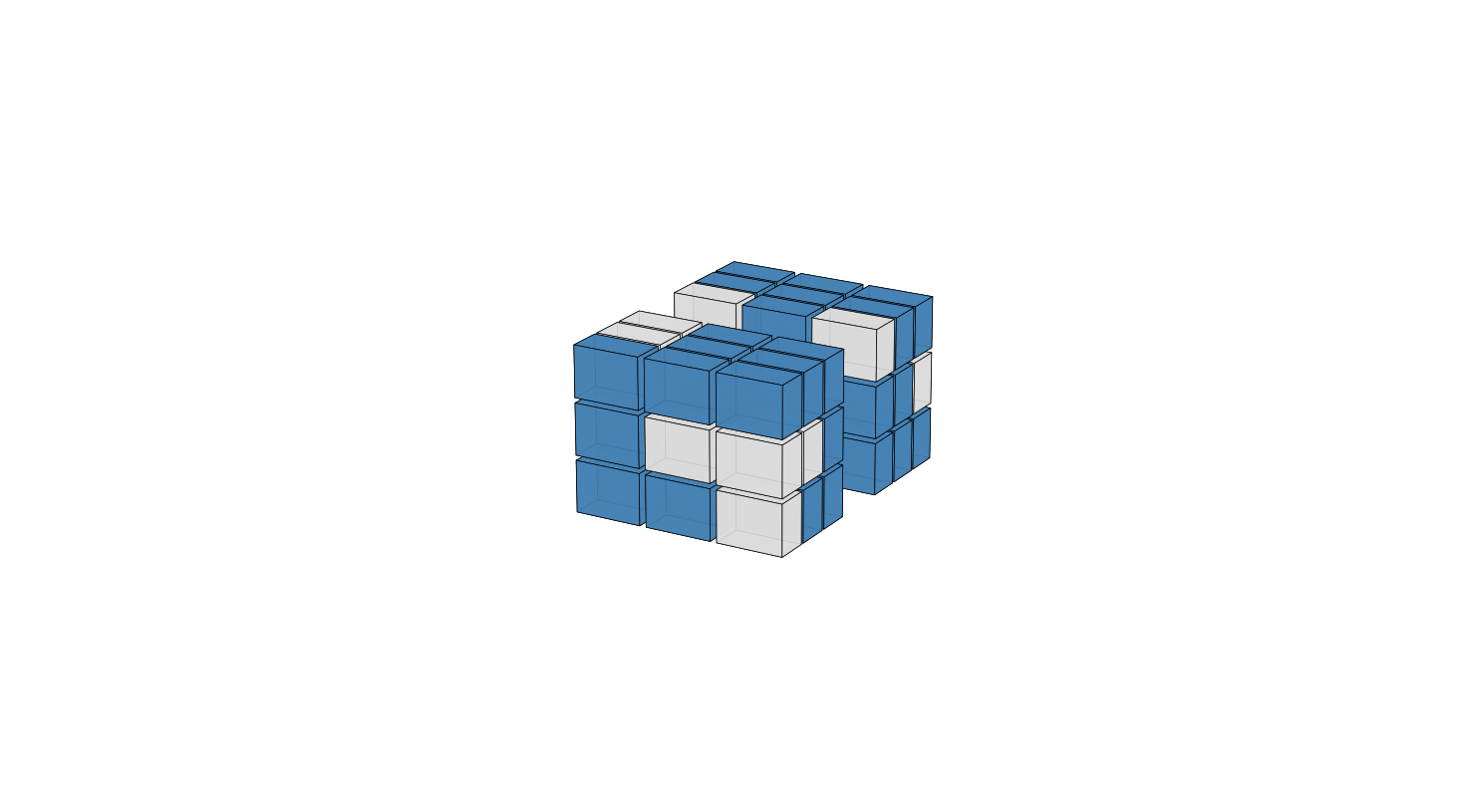} \hspace{5ex}
\includegraphics[height=.2\linewidth,trim=400 180 380 180,clip]{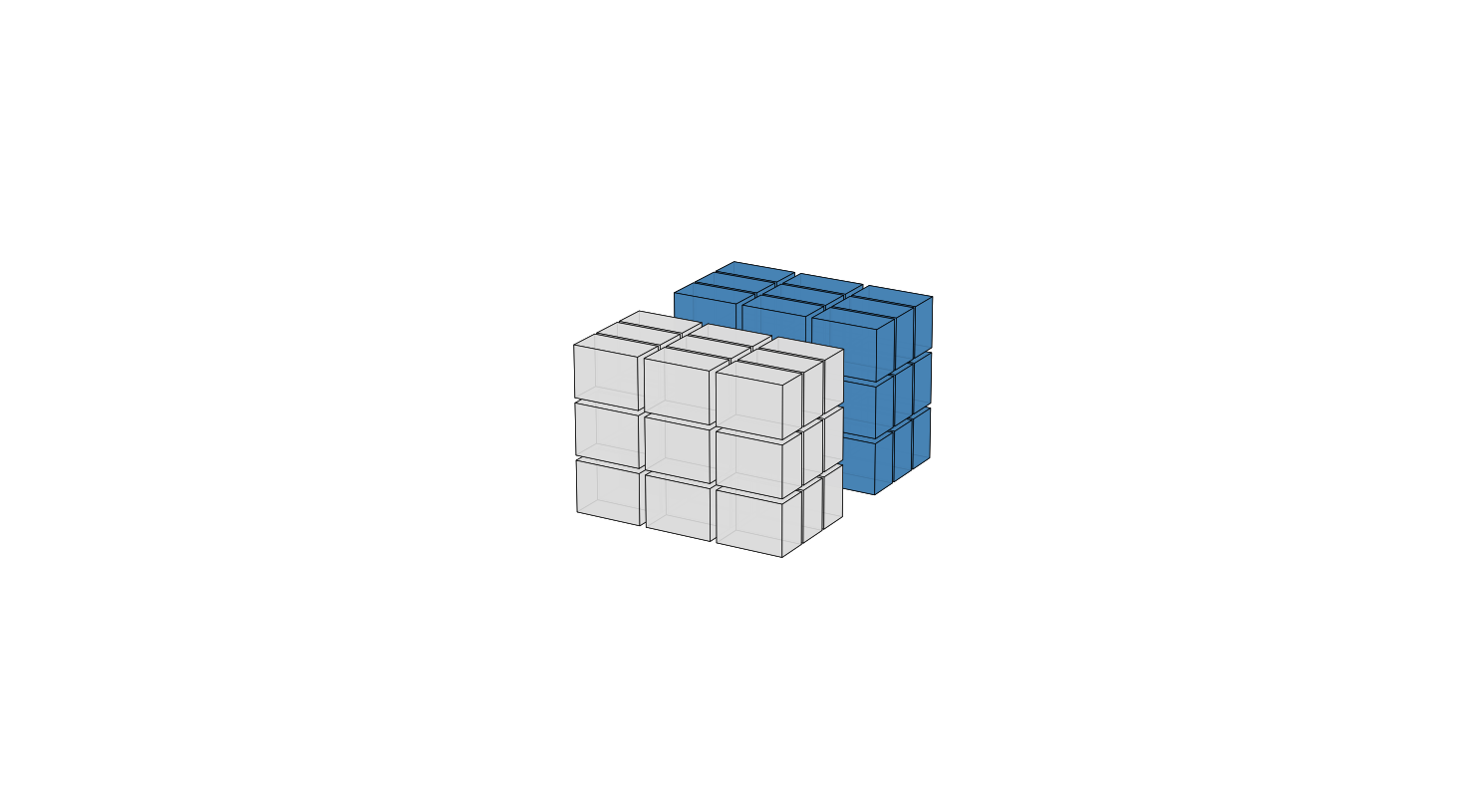}
\end{tabular}
\vspace{2pt}
\caption{Visualization of pruning strategies on two convolution channels with 3x3 filters and 3 channel input: dense (top), unstructured parameter pruning (left), and structured channel pruning (right).}
\label{fig:pruning_overview}
\end{figure}

Pruning can be performed in a structured or unstructured manner, as illustrated in \Cref{fig:pruning_overview} for a convolution layer. 
Unstructured pruning removes weights (connections/synapses) thus changing the NN structure, while structured pruning removes neurons (equivalently, convolutions/channels in CNNs) hence preserving the overall structure.   
Note that structured pruning of a neuron directly reduces the corresponding computation time and memory footprint for activations, whereas unstructured removal of weights may still require the computation of corresponding convolutions (with some zero filter weights), thereby not necessarily reducing computation or storage -- unless the utilized low-level libraries have support for related sparse operations.

\subsection{Gradual Channel Pruning}
Some pruning methods wait until the convergence of training to make any pruning decisions, since such a state provides certain theoretical properties, \eg gradients being negligible.
An iterative application of such pruning, however, requires huge resources, since the network needs to be retrained after each pruning step.
Alternatively, \emph{gradual pruning} aims to compress networks during training, thus learning simultaneously both the NN structure and the weights. 
For instance, GraNet~\cite{GraNet} and FGGP~\cite{FGGP} rank gradients and weight magnitudes to prioritize parameters to prune. 
Structured pruning of convolution channels in a gradual fashion is known as gradual channel pruning. 
This can use different pruning criteria such as LASSO regression~\cite{channelwise_pruning_He2017ChannelPF}, discrimination-aware loss~\cite{Discrimination_aware_channel_pruning}, or mean activation magnitude~\cite{stamp_paper}. 

Simultaneous Training and Model Pruning (STAMP)~\cite{stamp_paper} is a gradual channel pruning method specifically for UNet. 
It alternately trains for a fixed number of (so-called, \emph{recovery}) epochs and then removes the channel with the lowest normalized activations. 
For robustness of training to the disappearing channels, STAMP employs a targeted channel-wise dropout strategy with higher probabilities for channels that are more likely to be pruned soon.   
Its algorithmic prototype is shown in \Cref{alg:stamp}, as also used as a baseline in our work.
\begin{algorithm}
\caption{Simultaneous Training and Model Pruning (STAMP)~\cite{stamp_paper} for compressing Unet.}
\label{alg:stamp}
\begin{flushleft}
\textbf{while} all layers have at least one channel: \\
\hspace*{1.2em} Train for \textbf{recovery epochs} \\
\hspace*{1.2em} Rank filters by \emph{Pruning Criterion} \\
\hspace*{1.2em} Prune (remove) the lowest-ranked channel \\
\hspace*{1.2em} Update dropout probabilities \\
\textbf{end while}
\end{flushleft}
\end{algorithm}

\subsection{Unet architectures}
Inspired by encoder-decoder architectures, typical Unet~\cite{unet_paper} doubles feature channels at each encoder downsampling stage in order to preserve information content.
This principle is also inherited in any newer versions and improvements such as nnUnet~\cite{nnunet_isensee2021nnu} and Unet++~\cite{unet++}.
Such doubling with depth is indeed a primary factor inflating the total Unet parameter sizes.
Nevertheless, a key differentiator of Unet from simple encoder-decoder structures is the skip connections.
Since these enable direct information flow from encoders to decoders at corresponding levels, it can be argued whether channel doubling is really essential in a Unet architecture.
In this work, we study and test this hypothesis based on observations on gradual channel pruning and its variants.

\subsection{A Lean Unet (LUnet) architecture}
As seen later in our results, we test switching the STAMP pruning to a random version while keeping its architecture, as well as preferentially pruning the largest blocks (layers) regardless of their weights.
Based on our findings, we conclude that, as a general rule of thumb, minimized Unet architectures for segmentation can function successfully with channel sizes equal across blocks, especially at coarser levels since information preservation is not a concern thanks to skip connections. 
This leads us to the idea of using the limit case of such pruning, \ie a reduced Unet substructure having a fixed number of channels at every block/level.  
We propose this structure as a Lean Unet (LUnet), which can be seen as a new architecture or as a zero-shot pruning of standard Unet.
LUnet directly provides a small model without the overhead of pruning and involves no data-specific adaptation, both of which make it much less likely to overfit data.

\subsection{Evaluation and Datasets}
Analyses and evaluations were conducted on three datasets:
Hippocampus segmentation in MRI with a harmonized protocol (Harp)~\cite{frisoni2015harp}, which allows us to compare the results with baselines on a public dataset. 
Two internal datasets for submandibular gland (SG) and tracheal tree (TT) segmentation, respectively, complement this with evidence on CT images, with the latter dataset also providing a multi-label segmentation setting.
Dataset details are listed in \Cref{table:hyperparameters_flipped}.
We trained baseline Unet structures conventionally from initial random weights.
We also gradually pruned them during this same process using different strategies described later in the experiments.
We ran such gradual pruning to extreme sparsity until model collapse and reported the maximum Dice values observed during training.
This was chosen both to be consistent with baseline STAMP reporting and to show baselines in this best-case scenario.
For evaluation, we used the Dice similarity metric and average Dice across labels for the TT dataset.

\begin{table*}
\centering
\caption{Summary of dataset properties and hyper-parameters used.}
\label{table:hyperparameters_flipped}
\setlength{\tabcolsep}{4pt}
\renewcommand{\arraystretch}{1.1}
\begin{tabularx}{0.8\linewidth}{lccc}
\toprule
\textbf{Dataset} & \textbf{HarP} & \parbox[c]{.2\textwidth}{\centering \textbf{Submandibular Gland (SG)}} & \textbf{Tracheal Tree (TT)} \\
\midrule
Modality & T1 MRI & CT & CT \\
Image size (voxels) & 64x64x64 & 128x128x128 &  128x128x128 \\
Task (labels) & Hippocampus  & \parbox[c]{.2\textwidth}{\centering Submandibular Gland} & \parbox[c]{.2\textwidth}{\centering Trachea, Trachea extension,\\ Carina, Bronchus L/R}\\
\midrule
Train$\rightarrow$Test Splits & 200$\rightarrow$70 ; 50$\rightarrow$220 & 222$\rightarrow$56 & 64$\rightarrow$16 \\
Batch size & 16 & 1 & 1 \\
Recovery epochs & 5 & 1 & 1 \\
Levels in Unet & 5 & 4 & 4 \\
Convolutions per block & 2 & 3 & 3 \\
Channels-per-block at top layer & 4 & 24 & 24 \\
Learning rate & 0.01 & 0.001 & 0.001 \\
\bottomrule
\end{tabularx}
\end{table*}

\subsection{Implementation}

For STAMP\cite{stamp_paper} we used the code\footnote{\url{https://github.com/nkdinsdale/STAMP}} provided by its authors, by fixing a memory leak that initially prevented us from training on large images and longer training episodes.
We made minor modifications for compatibility with Python\,3.11 and PyTorch\,2.2, without altering any functionality. 
For comparability, the hyper-parameters for the HarP dataset were chosen to closely replicate the STAMP settings~\cite{stamp_paper} with a relatively low number of convolution filters in the top (initial) Unet level, which enabled larger batch sizes.
For the CT datasets TT and SG, we employed a much larger 3D Unet baseline architecture that is representative of typical Unet model sizes used in production and in the literature.
This allowed a batch size of only one image volume to fit in memory.
For the CT datasets, preliminary experiments indicated that a recovery epoch setting of one was sufficient for results comparable to higher settings. 
A summary of varying hyper-parameters is provided in \Cref{table:hyperparameters_flipped}.

We used L$_2$-norm of channel activations as the pruning criterion as suggested in STAMP, and a base probability of 5\% for targeted dropout.
All models were trained using the ADAM optimizer and momentum parameters of $\beta_1$=0.9 and $\beta_2$=0.999. 
Due to the smaller batch sizes of SG and TT, a relatively smaller learning rate was needed for these.
All experiments were run on an NVIDIA RTX A5000 GPU with 24 GB of VRAM.

\section{Experiments and Results}
On the three datasets, we conducted experiments under 4 settings, where HarP was tested with two different train-test splits, see \Cref{table:hyperparameters_flipped}.
Each experiment was repeated three times using random initializations, reporting their median and MAD (median absolute deviation). 
In the interest of space, we present the analysis figures below only for HarP 200$\rightarrow$70 split, with supplementary analysis figures for 50$\rightarrow$220 split provided later in \Cref{fig:supp_harp50} in Appendix.

\subsection{Gradual Channel Pruning}
\begin{figure}
    \centering    
    \resizebox{\linewidth}{!}{\input{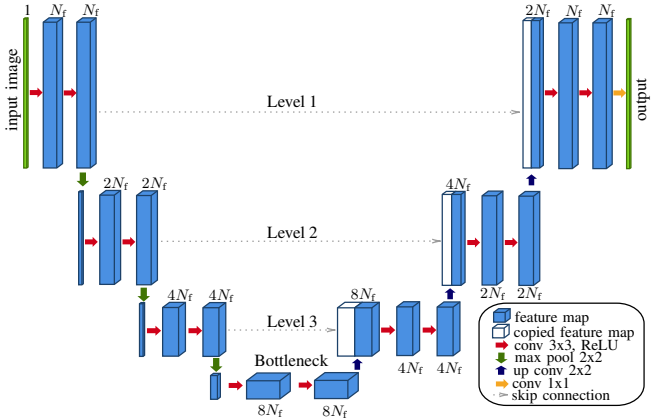}}
    \caption{Visualization of a regular Unet architecture, with $N_\text{f}$ starting convolution filters, two convolutions per block, and a depth of four levels. LUnet architecture follows the same structure, except that the number of convolution channels per block remains constant across network levels rather than doubling at each successive level.
    }
    \label{fig:unet}
\end{figure}

\begin{figure*}
\begin{tabular}{@{}cc@{}}
\includegraphics[height=.3\textwidth,width=.45\textwidth]{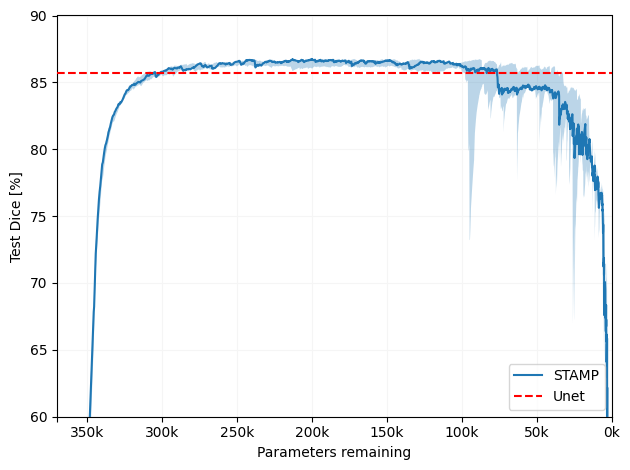} &
\includegraphics[height=.3\textwidth,width=.47\textwidth]{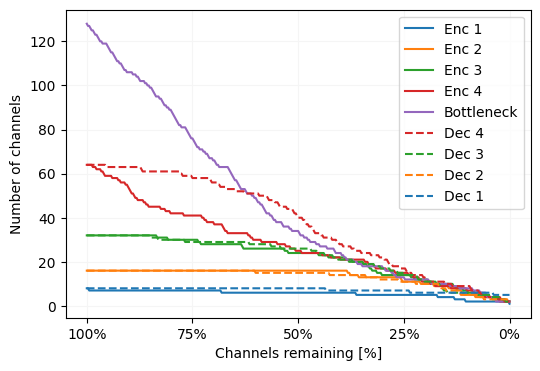} \\[-1ex]
\ss{(a) Pruning vs. dense Unet} & \ss{(b) Evolution of channels per layer}\\[1ex]
\multicolumn{2}{c}{\includegraphics[width=.99\textwidth]
{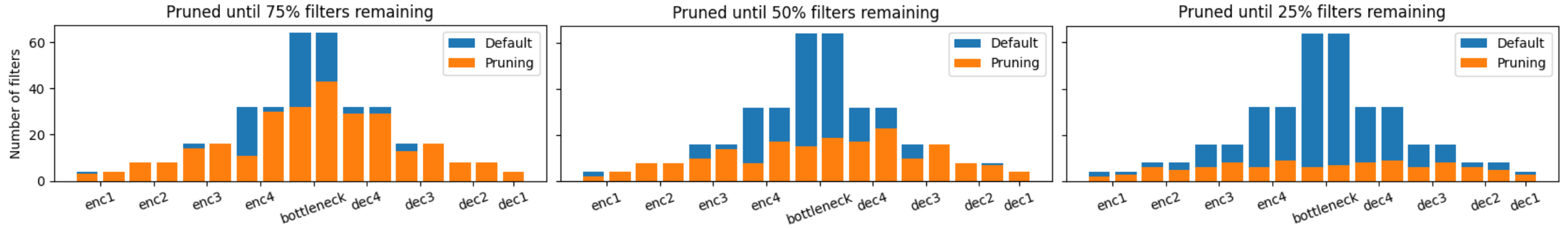}}\\[-1ex]
\multicolumn{2}{c}{\ss{(c) Distribution of channels per block}}
\end{tabular}
\caption{Pruning on HarP for 200$\rightarrow$70 train-test split. (a)~Test Dice score evaluated during training with STAMP set compared with the dense Unet baseline (Unet$_{100\%}$).
(b)~Remaining channels at each Unet encoder-decoder subpart during pruning. 
(c)~Distribution of channels per block at three sample pruning steps when \{75,50,25\}\% of total channels remain.
}
\label{fig:pruning}
\end{figure*}

We first trained a dense Unet baseline with an initial block size $N_\mathrm{f}$=4 filters, with an architecture seen in \Cref{fig:unet}.
We then gradually pruned its channels using STAMP.
Their comparative results seen in \Cref{fig:pruning}(a) corroborate the results presented in~\cite{stamp_paper}, where pruning is seen to outperform the Unet baseline across a wide range of sparsity values.

Analyzing channel sizes at each encoder-decoder layer as seen in \Cref{fig:pruning}(b), Unet components with the largest channel counts are observed to be preferentially pruned first.
Once the bottleneck and deeper layers are pruned down to channel sizes of shallower layers, only then the latter start being pruned as well.
This causes the Unet architecture to slowly become flatter in channel size, starting with the deepest layers equalizing first.
This pattern can also be observed in \Cref{fig:pruning}(c) which shows snapshots of individual channel sizes at each Unet block at three different pruning stages that correspond to when \{75,50,25\}\% of total channels remain.
This observation may be explained by the synaptic saliency~\cite{SynFlow} of a channel at a wider block or layer being on average lower, hence making those channels more likely to be selected by a chosen pruning criterion.
This observation begs the important question: is the architecture attained by pruning more pertinent than the actual channels selected for pruning?
This somewhat parallels the arguments in~\cite{Rethinking_the_value_of_nn_pruning_Liu19} questioning the Lottery Ticket Hypothesis~\cite{lottery_ticket_hyp_frankle2018lottery}.

\subsection{Compressed architecture and parameter values}
\begin{figure*}
\centering
\begin{tabular}{c|c}
\includegraphics[height=.35\textwidth]{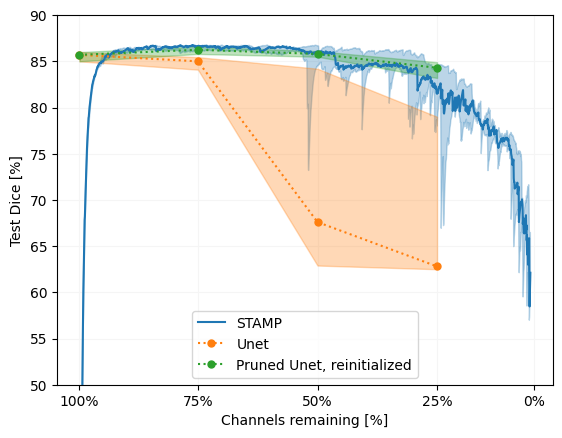} & 
\includegraphics[height=.345\textwidth]{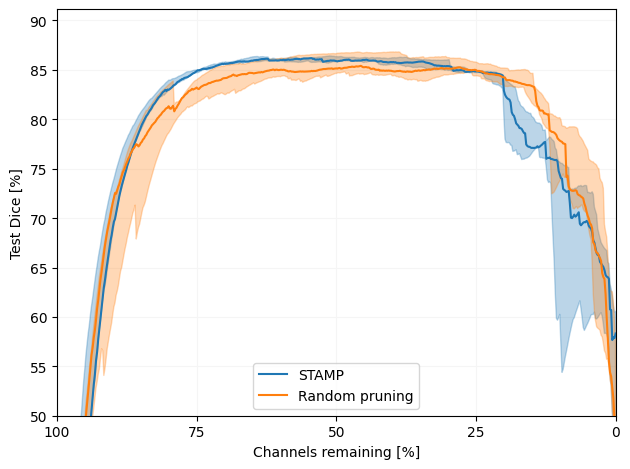}\\[-1ex]
\ss{(a) Dense vs. scaled-down vs. reinitialized-pruned Unets} & \ss{(b) Random pruning from STAMP-chosen block. ($N_\mathrm{f}$=8, 1 recovery epoch)}\\
\end{tabular}
\caption{Dice comparisons of strategies.
(a)~Pruning compared to the baseline dense Unet$_{100\%}$ and linearly-scaled-down models Unet$_{\{75\%,50\%,25\%\}}$ (orange) as well as equivalent-sized pruned networks after retraining from random initializations (green).
(b)~Pruning a random channel from STAMP-determined block (orange), compared to pruning based on channel activations (blue).
Curves show the median values, with min/max ranges of three repeated runs shaded.}
\label{fig:architecture}
\end{figure*}

To test the above hypothesis, we took the network structures pruned above at \{75,50,25\}\% channels-remaining and retrained these from scratch by reinitializing their parameters with random values.
These are seen in \Cref{fig:architecture}(a) being non-inferior to STAMP pruning -- and even superior at a high sparsity of 25\%.
Hence, in this particular setting, the network architecture seems to be more important than the actual weights during pruning.
To check whether this advantage is simply due to less overfitting of a smaller network, we also tested corresponding Unet baselines (called Unet$_{\{75\%,50\%,25\%\}}$) whose number of parameters were scaled linearly at all blocks.
These baselines, however, performed significantly worse than both pruning and pruned-and-reinitialized models, especially at higher sparsity.
This indicates that the location of channels are indeed important, not merely the number of channels in total.

\begin{figure*}
\begin{tabular}{@{}c@{\,}c@{\,}c@{}}

\includegraphics[clip,trim=0 0 40 40,height=.295\textwidth,width=.3\textwidth]{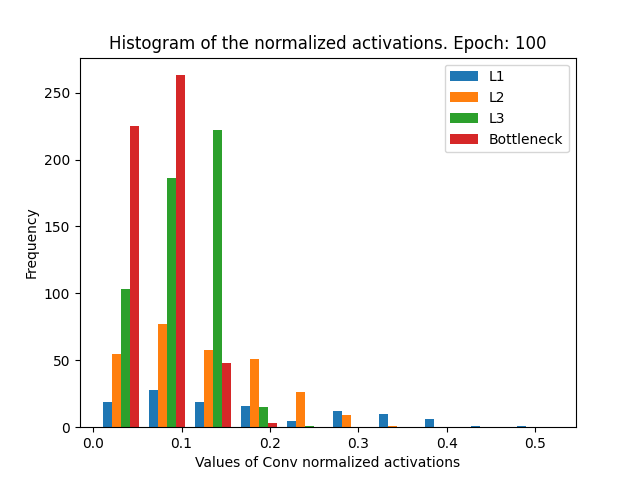} &
\includegraphics[height=.3\textwidth,width=.35\textwidth]{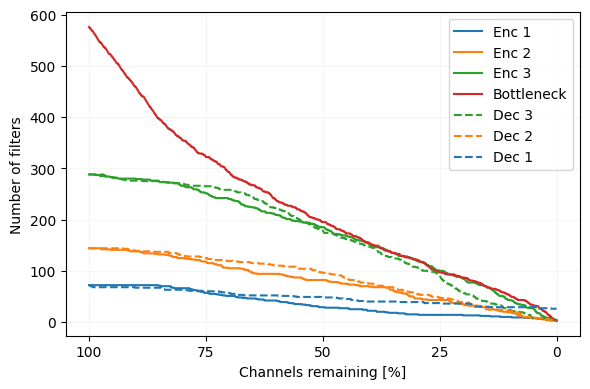} & 
\includegraphics[height=.3\textwidth,width=.35\textwidth]
{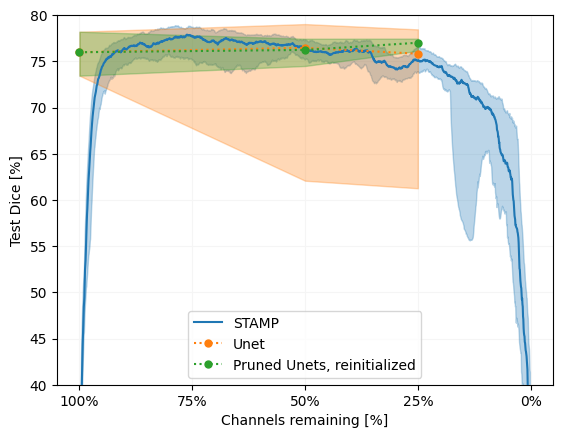}\\[-1ex]
\ss{(a) Histogram of criterion} & \ss{(b) Evolution of channels per layer} & \ss{(c) Reinitialized pruned Unets}\\[1ex]
\multicolumn{3}{c}{\includegraphics[width=\textwidth]
{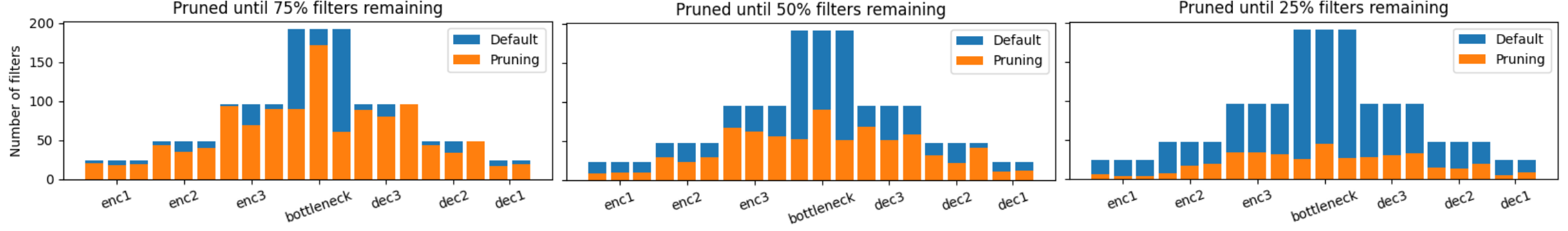}}\\[-1ex]
\multicolumn{3}{c}{\ss{(d) Distribution of channels per block}}
\end{tabular}
\caption{Sample results on the Tracheal Tree (TT) dataset: 
(a)~Distribution of pruning criterion (normalized activations) per Unet level at 95\% channels remaining. (b)~Remaining channels at each Unet encoder-decoder subpart during pruning. 
(c)~Pruning compared to the baseline dense and linearly-scaled-down Unet$_{\{100\%,75\%,50\%,25\%\}}$ (orange) as well as equivalent-sized pruned networks after retraining from random initializations (green).
(d)~Distribution of channels per block at the pruning steps when \{75,50,25\}\% of total channels remain.
}
\label{fig:TTresults}
\end{figure*}

Since the channels of a block are symmetric and interchangeable considering random reinitializations in the above experiments, what is important must then be the overall architecture achieved by pruning, not the exact channels selected during it.
We tested this hypothesis by pruning a random channel from the pruning-identified network block, \ie not necessarily the exact channel chosen by the STAMP pruning criterion.
In other words, we ran the pruning as normal and let it choose a channel, but then instead of removing that channel we removed a randomly-sampled channel within that same block.
This effectively achieves the same architecture and topology evolution as in \Cref{fig:pruning}(b) but with exact channels not being based on the pruning criterion.
\Cref{fig:architecture}(b) shows that such random pruning from the given block achieves similarly to or even better than targeted channel pruning of STAMP, confirming that the pruned architecture is the key -- not necessarily the weights.
Some corresponding results for the TT dataset can be seen in \Cref{fig:TTresults}.

\subsection{Systematic widest-channel pruning}
\begin{figure*}
\centering
\begin{tabular}{cc}
\includegraphics[height=0.32\textwidth]{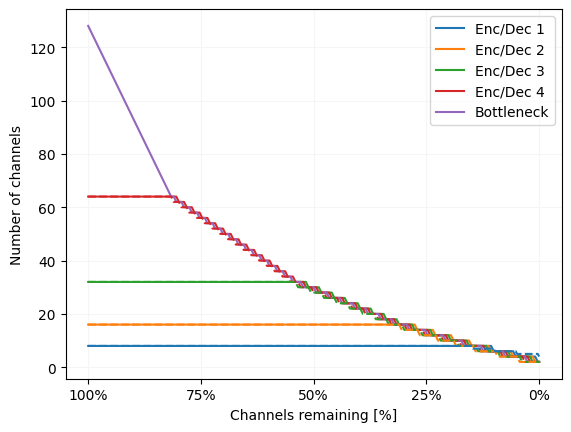} &
\includegraphics[height=0.32\textwidth]{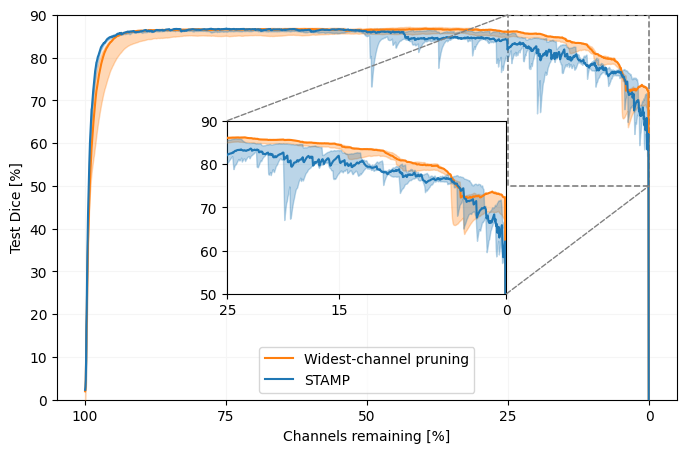}\\[-1ex]
\ss{(a) Prescribed layer sparsity for widest-block pruning} & \ss{(b) Comparison of Widest-block and criterion-based pruning}\\
\end{tabular}
\caption{(a)~Prescribed number of channels per Unet subpart for pruning always from the block with most channels, \ie \emph{widest-block pruning}. (b)~Dice evolution of widest-block pruning and STAMP.}
\label{fig:widest}
\end{figure*}

To achieve flatter architectures similar to those observed above in \Cref{fig:pruning}(b-c) but without using any pruning criterion, we next tested a na\"ive approach of pruning a random channel from the largest block at each step.
This results in a preset pruning schedule seen in \Cref{fig:widest}(a) that is independent of training data and takes a direct shortcut to a flat channel distribution.
Strikingly, this simple approach outperforms standard STAMP pruning, especially at extreme sparsity values, as seen in \Cref{fig:widest}(c).

\subsection{Lean Unet}
STAMP approximating a flat architecture and the above simplified widest-channel pruning performing similarly or superiorly to criterion-based STAMP lead us to our proposed solution Lean Unet (LUnet) with a fixed channel size in all blocks.
Accordingly, the starting (top-level) channel size stays the same across all layers.
LUnet results are tabulated in \Cref{table:results_harp_compact} comparatively with the variations presented above, for both train-test splits of HarP dataset.
A comparison of LUnet and its baselines for the SG and TT dataset can be seen in \Cref{table:results_SG-TT_compact}.

\begin{table*}
\caption{Median results ($\pm$ median absolute deviation) for different Unets on the HarP dataset. Columns indicate $N_\text{ch}$ the total number of channels in the architecture, $N_\text{p}$ the total number of network parameters, and $N_\text{f}$ the number of channels per block in the top Unet level ($f$~in~\cite{stamp_paper}).
Pruning may reduce $N_\text{f}$, so its value at initialization is indicated in parentheses ($\cdot$).}
\label{table:results_harp_compact}
\begin{tabular}{@{}c@{}}
\includegraphics[width=\textwidth]{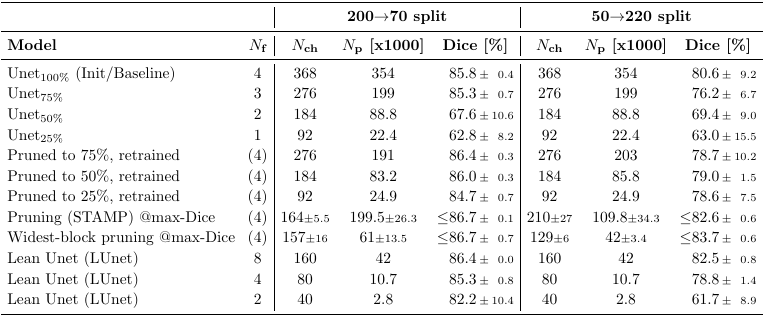}
\end{tabular}
\end{table*}

\begin{table*}
\caption{Median results ($\pm$ median absolute deviation) for different Unets on the Submandibular Gland (left) and Tracheal Tree (right) datasets. Columns indicate  $N_\text{ch}$ the total number of channels in the architecture, $N_\text{p}$ the total number of network parameters, and $N_\text{f}$ the number of channels per block in the top Unet level ($f$~in~\cite{stamp_paper}).
Pruning may reduce $N_\text{f}$, so its value at initialization is indicated in parentheses ($\cdot$).} 
\label{table:results_SG-TT_compact}
\begin{tabular}{@{}c@{}}
\includegraphics[width=\textwidth]{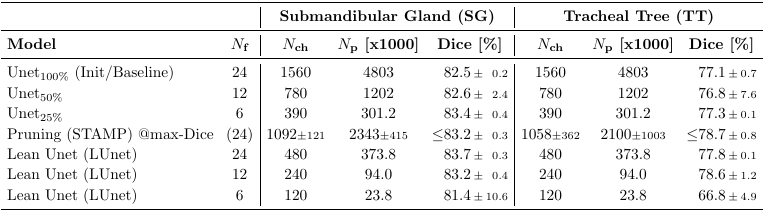}
\end{tabular}
\end{table*}

For pruning, performance is reported at the sparsity where it achieves its highest test Dice score, which is a highly biased metric indicating a best-case scenario unattainable in practice.
Furthermore, the large deviation in $N_\mathrm{p}$ for these values indicate that the maximum Dice is achieved at different model sizes and hence pruning stages, which cannot be controlled nor predicted apriori, also indicating a lack of repeatability.
Our results show that the proposed LUnet structure achieves similar performance, while being a fixed and data-independent structure, without requiring costly pruning operations to determine.
With parameter counts much smaller than STAMP, LUnet can achieve similar performance; \eg 199 vs. 42\,K parameters for HarP in \Cref{table:results_harp_compact}.
Moreover, conventional Unet architecture with linear scaling is seen to collapse around 88.8\,K parameters (Unet$_{50\%}$) while our LUnet performs well even with a fraction of the parameter count (\eg 2.8\,K).
With merely 42\,K parameters, LUnet outperforms the largest Unet baseline (Unet$_{100\%}$) with 354\,K parameters.
Indeed, in the HarP setting for a fixed top-level block size $N_\mathrm{f}$=4 channels, Unet contains 354/10.7$>$33 times more parameters than LUnet, and similarly 88.8/2.8$>$31 times more for $N_\mathrm{f}$=2.

\section{Discussion and Conclusions}
By analyzing the distribution of channels during pruning and reinitializing pruned networks, we have shown that (1)~pruning primarily removes channels in deeper layers and (2)~reinitialized pruned models attain comparable performance when retrained.
These insights motivated two simplified pruning strategies, both targeting the deeper layers first by pruning (1)~some random channel from the block selected based on pruning criterion and (2) removing a random channel from the widest block at a given time. 
Both approaches achieve performance comparable or superior to original pruning, suggesting that pruning effectiveness depends more on the Unet structure it attains than on the specific channels it retains.
We have proposed \emph{LUnet}, a compact Unet variant with fewer channels in the bottleneck and deeper layers. 
LUnet matches the performance of Unet baselines while using over 30$\times$ fewer parameters.
This result challenges the long-standing assumption that the number of channels should double after each downsampling -- a design convention that has persisted since the original Unet architecture.
With its much smaller structures, LUnet also matches STAMP performance as reported at its maximum test value -- a reporting convention that we inherited from its original publication~\cite{stamp_paper}.
Note that such reporting of self-maximized test score is largely dependent on noise/stochasticity in training iterations and also represents a highly biased upper-bound scenario, which is not likely to generalize on a separate hold-out data; hence we prepended these values with a "$\leq$" in the tables.
This is corroborated by the STAMP result reporting being an upper-bound for the retrained models at \{25, 50, 75\}\% sparsity.
For all models trained without pruning, such as LUnet, the reported test Dice are unbiased to test set, since the models and weights are fixed once the training ends.
Therefore, comparisons with pruned and retrained models are more representative of relative performance gains in real-world scenarios.
Furthermore, pruning involves considerable computational overhead due to repeated model architecture updates and any additional recovery epochs. 
LUnet, in contrast, achieves competitive results with a fraction of the network size and in a data-independent, robust manner.

\bibliographystyle{IEEEtran}
\bibliography{references}
\newpage
\appendix
\begin{figure}[h]
\centering
\begin{tabular}{c}
\includegraphics[height=.3\textwidth]{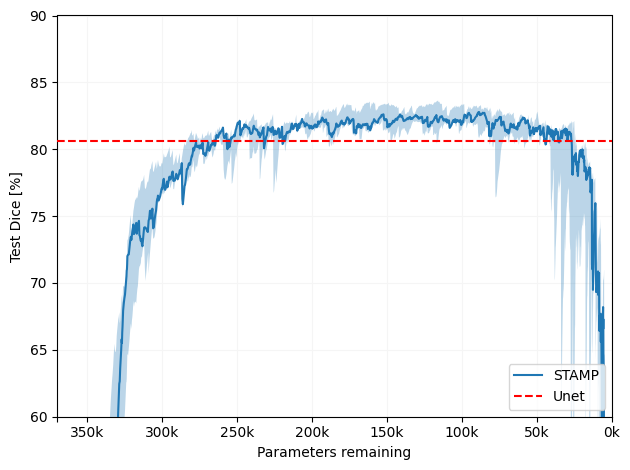} \\[-1ex]
\ss{(a) Pruning vs. dense Unet} \\[1ex]
\includegraphics[height=.3\textwidth]{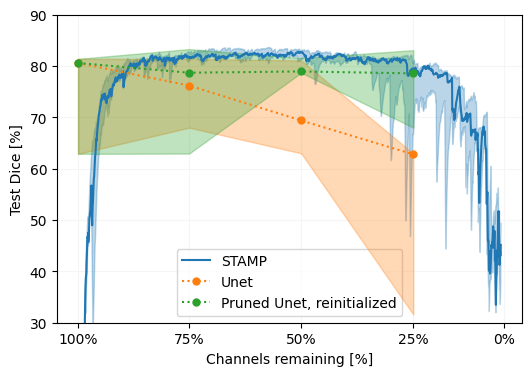}\\[-1ex]
\ss{(b) Dense vs. scaled-down vs. reinitialized-pruned}\\[1ex]
\includegraphics[height=.3\textwidth]{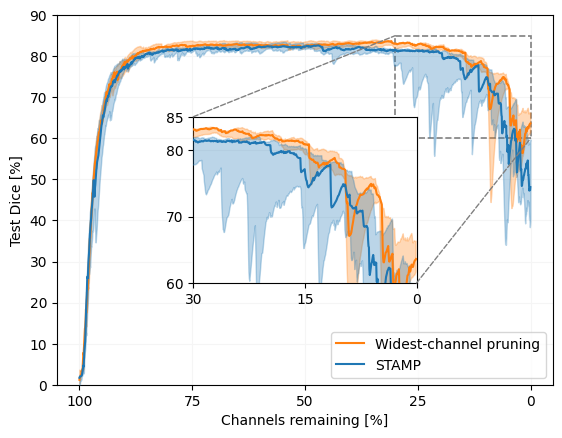}\\[-1ex]
\ss{(c) Dice evolution of widest-block pruning and STAMP.}\\[1ex]
\end{tabular}
\caption{Result figures on HarP with 50$\rightarrow$220 train-test split. (a)~Pruning compared to the baseline Unet. (b)~Pruning compared to the baseline dense and linearly-scaled-down Unet$_{\{100\%,75\%,50\%,25\%\}}$ (orange) as well as equivalent-sized pruned networks after retraining from random initializations (green). (c)~Dice evolution of STAMP based on channel activations compared to widest-block pruning. 
Curves show the median values, with min/max ranges of three repeated runs shaded.}
\label{fig:supp_harp50}
\end{figure}
\end{document}